\documentclass{article}

\usepackage{microtype}
\usepackage{graphicx}
\usepackage{booktabs}
\usepackage{multirow}
\usepackage{array}
\usepackage[table]{xcolor}
\usepackage{hyperref}

\usepackage[preprint]{icml2026}

\usepackage{amsmath,amssymb,mathtools}
\usepackage{algorithm}
\usepackage{algorithmic}
\usepackage{tikz}
\usepackage{pgfplots}
\usetikzlibrary{arrows.meta,positioning}
\pgfplotsset{compat=1.18}

\definecolor{methodblue}{RGB}{44,105,182}
\definecolor{solvergreen}{RGB}{0,145,116}
\definecolor{harnessorange}{RGB}{230,150,55}
\definecolor{hzeropurple}{RGB}{132,92,191}
\definecolor{bestshade}{RGB}{255,232,224}
\definecolor{secondshade}{RGB}{230,239,255}
\newcommand{\best}[1]{\cellcolor{bestshade}\textbf{#1}}
\newcommand{\second}[1]{\cellcolor{secondshade}\underline{#1}}

\newcommand{\method}{\mbox{JoMoHa}}
\newcommand{\engineer}{\mathrm{E}}
\newcommand{\solver}{\mathrm{S}}
\newcommand{\E}{\mathbb{E}}

\newcommand{\noharness}{\varnothing}

\icmltitlerunning{JoMoHa: Joint Model--Harness Optimization with Reinforcement Learning}

\begin{document}

\twocolumn[
\icmltitle{JoMoHa: Joint Model--Harness Optimization
with Reinforcement Learning}
\begin{icmlauthorlist}
  \icmlauthor{Xudong Wu}{hku}
  \icmlauthor{Guangchen Lan}{sales}
  \icmlauthor{Mingqi Yuan}{hku}
  \icmlauthor{Jiayu Chen}{hku}
\end{icmlauthorlist}

\icmlaffiliation{hku}{The University of Hong Kong, Hong Kong}
\icmlaffiliation{sales}{Salesforce AI Research}

\icmlcorrespondingauthor{Jiayu Chen}{jiayuc@hku.hk}
\icmlkeywords{multi-agent reinforcement learning, language agents,
joint harness--weight optimization}
\vskip 0.3in
]
\printAffiliationsAndNotice{}

\begin{abstract}
The performance of language agents depends jointly on model
weights and the harness.
Optimizing either alone can be constrained by the other
component, limiting overall performance gains.
Our analysis of naive alternating training reveals progressive
incompatibility between the model and harness, which increasingly
interferes with task execution and may also undermine model
training, leading to performance degradation.
To coordinate the evolution of models and harnesses, we
introduce \method{}, to our knowledge the first framework
to jointly train an executable harness generation policy
and a task execution policy through reinforcement learning.
\method{} outperforms the evaluated weight-only and harness-only
baselines and achieves the highest success rates among the
evaluated methods on ALFWorld, WebShop, and DBBench,
averaging 70.3\% and reaching 97.0\% on ALFWorld.
By learning both task execution and runtime improvement
from its own generated trajectories, \method{} performs
recursive self-improvement (RSI).
\end{abstract}

\begin{figure}[!ht]
\centering
\begin{tikzpicture}
\begin{axis}[
  width=0.98\linewidth,height=4.8cm,
  xlabel={Reported training step},ylabel={Success (\%)},
  xmin=0,xmax=148,ymin=70,ymax=103,
  xtick={0,36,72,108,144},ytick={70,80,90,100},
  tick label style={font=\footnotesize},
  label style={font=\small},
  axis lines=left,axis line style={black!45},
  tick style={black!35},
  ymajorgrids=true,grid style={black!9},
  axis background/.style={fill=methodblue!2},
  legend columns=2,
  legend style={font=\footnotesize,at={(0.5,1.03)},
    anchor=south,draw=none,fill=none},
  legend cell align=left
]
\addplot[methodblue,line width=1.6pt,mark=*,mark size=2.4pt,
  mark options={draw=white,line width=0.45pt}] coordinates
  {(0,83.0) (36,87.0) (72,93.5) (108,93.5) (144,97.0)};
\addlegendentry{\method{}}
\addplot[solvergreen,line width=1.1pt,mark=square*,mark size=2pt,dashed,
  mark options={draw=white,line width=0.4pt}] coordinates
  {(0,73.5) (36,77.0) (72,80.5) (108,88.0) (144,89.5)};
\addlegendentry{Weight-only}
\addplot[harnessorange,line width=1.1pt,mark=triangle*,mark size=2.4pt,dashdotted,
  mark options={draw=white,line width=0.4pt}] coordinates
  {(0,83.0) (36,83.0) (72,82.5) (108,85.5) (144,80.5)};
\addlegendentry{Harness-only}
\addplot[hzeropurple,line width=1.1pt,mark=diamond*,mark size=2.5pt,dotted,
  mark options={draw=white,line width=0.4pt}] coordinates
  {(36,83.0) (72,79.0) (108,91.5) (144,90.5)};
\addlegendentry{Frozen Engineer}
\node[font=\footnotesize,methodblue,anchor=south east,xshift=-3pt]
  at (axis cs:144,97.0) {\textbf{97.0}};
\node[font=\footnotesize,hzeropurple,anchor=south east,xshift=-3pt]
  at (axis cs:144,90.5) {\underline{90.5}};
\end{axis}
\end{tikzpicture}
\caption{\textbf{Learning the runtime writer alongside the Solver on ALFWorld.}
The joint system reaches 97.0\%, versus 90.5\% for Frozen Engineer,
89.5\% for Weight-only, and 80.5\% for Harness-only.
Weight-only does not use harness.
Frozen Engineer keeps the initial code-writing model fixed but generates
harnesses while training the Solver.}
% Confidence intervals are reported in Appendix~\ref{app:statistics}.

\label{fig:curves}
\end{figure}
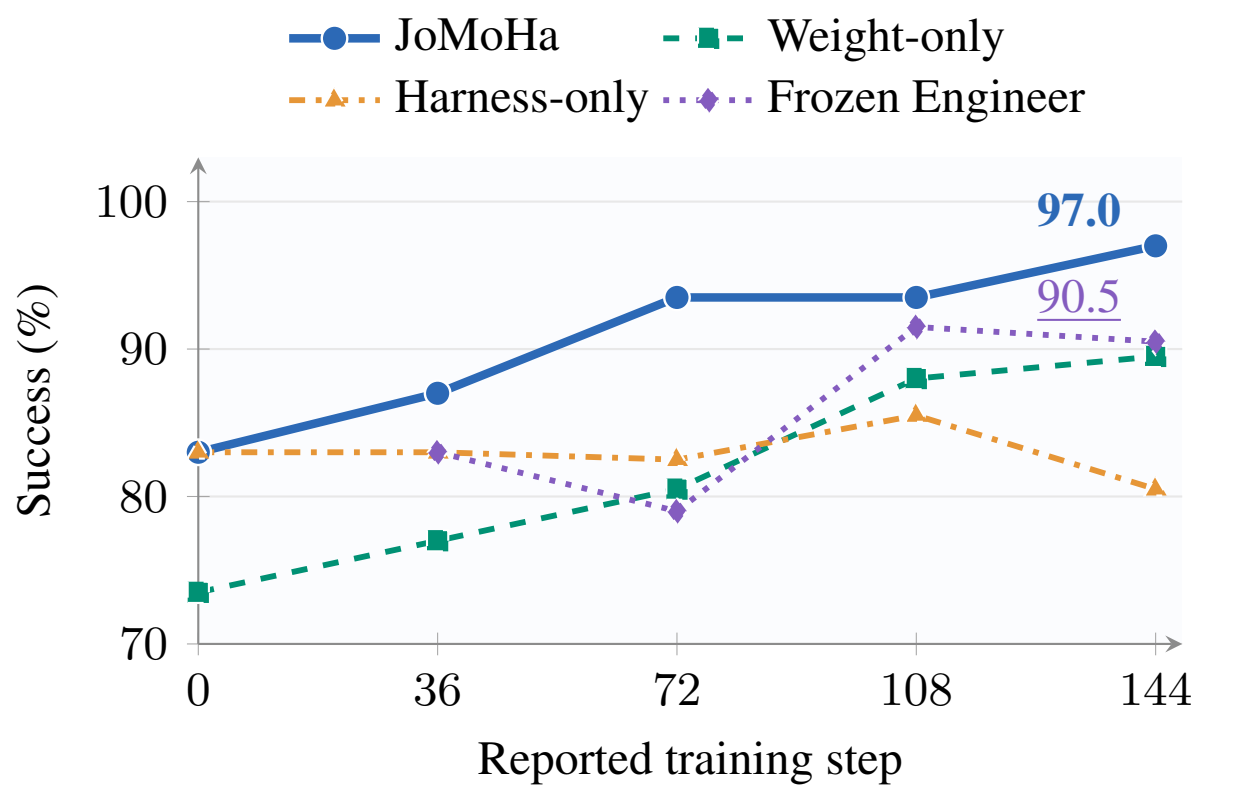

\begin{figure*}[t]
\centering
\includegraphics[width=\textwidth]{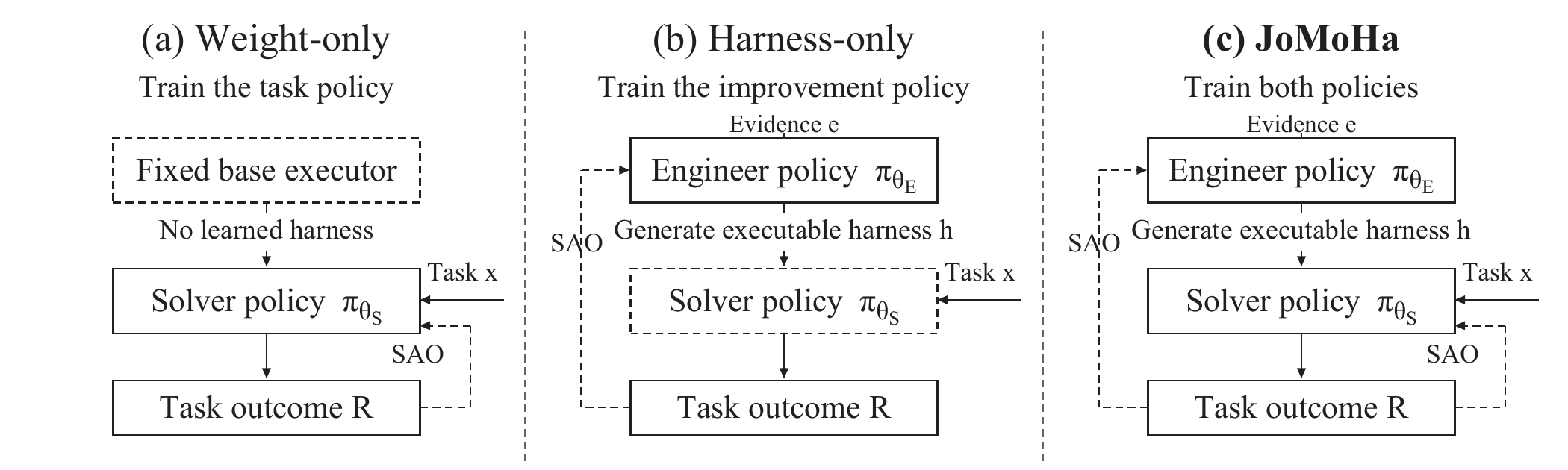}
\caption{\textbf{From one-sided learning to joint model--harness RL.}
Weight-only trains the Solver under the fixed base executor, without a
learned harness. Harness-only trains the Engineer against a frozen Solver.
\method{} trains both policies from the same executed task outcome.
Solid policy boxes denote updated parameters; dashed boxes denote fixed
components. Dashed arrows carry training feedback, not gradients through
the generated program. SAO denotes the single-rollout actor--critic
backbone \citep{hou2026sao}; Section~\ref{sec:background} introduces the learning terms and
notation.}
\label{fig:comparison}
\end{figure*}

\begin{figure*}[t]
  \centering
  \includegraphics[width=\textwidth]{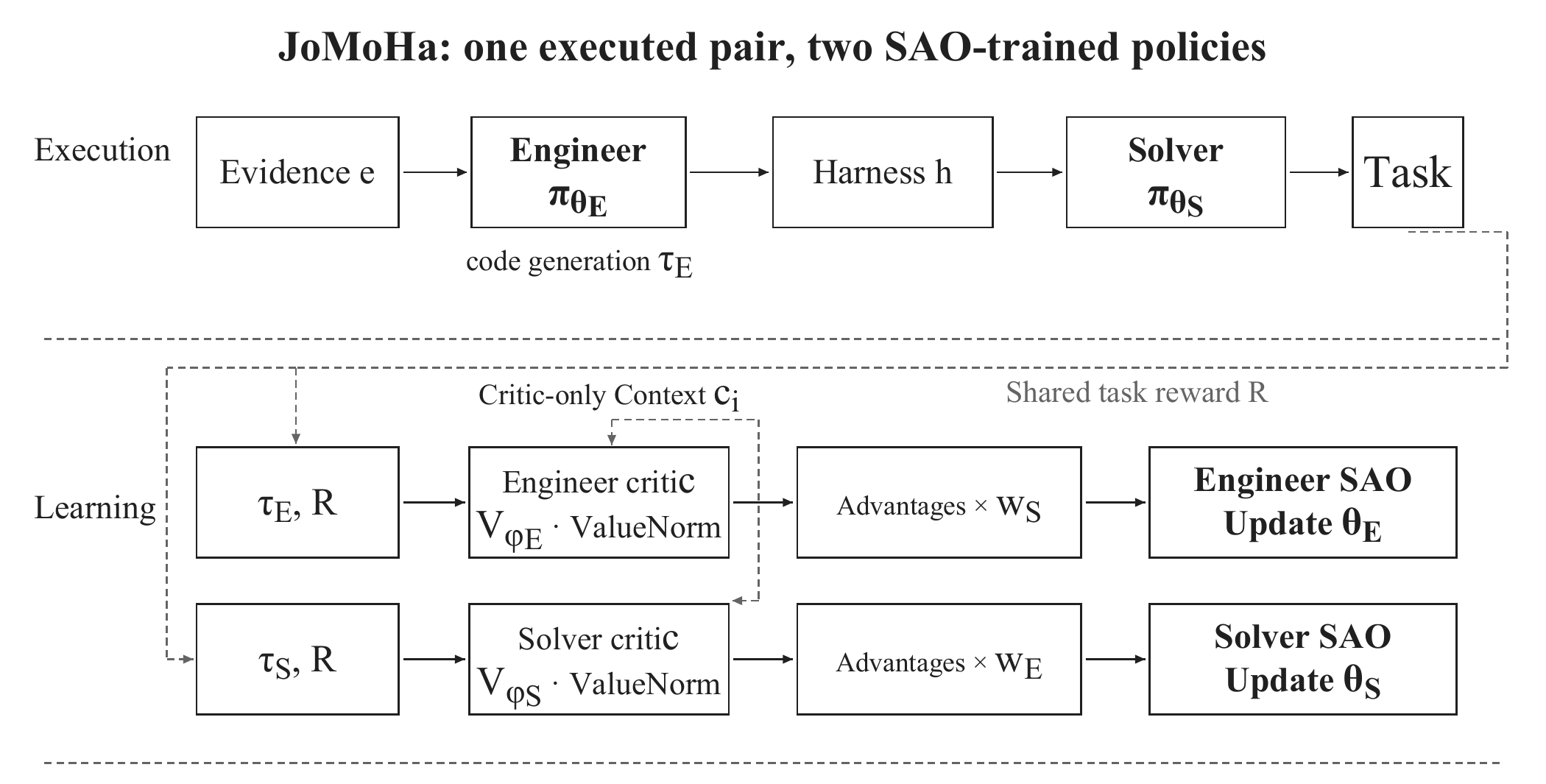}
  \caption{\textbf{Joint reinforcement learning of the Engineer and Solver.}
  The Engineer writes a runtime program under which the Solver interacts
  with the task. The resulting shared reward \(R\) trains both policies,
  each on its own generated tokens.}
  \label{fig:overview}
\end{figure*}
\section{Introduction}
\label{sec:intro}

A language agent consists of a model and a harness.
The model weights determine its learned reasoning and decision-making
capabilities, while the harness organizes how these capabilities are
used.
Improving an agent therefore admits two complementary paths:
training its model weights and improving its harness.

Existing methods have made progress along both paths.
Reinforcement learning methods train model weights from task feedback,
improving the policy that makes decisions
\citep{yu2022mappo,shao2024deepseekmath}.
Alternatively, harness optimization changes the runtime surrounding
the model \citep{lee2026metaharness,shao2026harnessr1}. However, agent performance depends jointly on both components,
and optimizing either alone can leave overall gains constrained
by the other.
The comparison with weight-only and harness-only training
in Figure~\ref{fig:curves} further motivates learning model
weights and harnesses together.

However, naively alternating updates to model weights and
the harness does not ensure their compatibility.
As the model changes, an existing harness may provide
outdated or incorrect guidance that interferes with task
execution and limits performance gains.
Such guidance also shapes the experience used for training
and may steer subsequent model updates toward less effective
behavior, potentially worsening performance over time.
We diagnose these issues through controlled comparisons
and failure trajectory analysis in Section~\ref{sec:problem}.

WHALE mitigates this issue through small alternating updates
to model weights and harnesses.
Its published instantiation updates model weights through
online rejection-sampling fine-tuning (RSFT) rather than RL,
and uses Meta-Harness search driven by a fixed external
proprietary frontier model \citep{kim2026whale}.
However, harness optimization continues to rely on a fixed
external proprietary frontier model to propose improvements.
Execution feedback guides harness selection, but does not
update the proposer itself.
The capability to improve the harness therefore remains
delegated to an external model rather than trained within
the system.
This motivates a different approach: making harness generation
a trainable policy and using the same task execution outcome
to train both the harness generator and the task performer.

\textbf{We instantiate recursive self-improvement (RSI)}
by learning both task execution and harness construction
from the system's own generated programs, trajectories,
and execution feedback.
Harness-R1 supports the potential of this direction:
its trained harness Engineer outperforms the evaluated fixed
frontier editors, including proprietary models
\citep{shao2026harnessr1}.
However, it keeps the task-performing model frozen during
Engineer training.
This leaves an open question:

\noindent\textbf{How can both policies improve through reinforcement
learning from their shared task executions?}

A natural starting point is to alternate updates to the two policies
using task outcomes.
Yet alternation alone does not ensure that their improvements remain
compatible.
We investigate this problem through controlled diagnosis of a
long-phase alternating run in which performance initially improves
and subsequently deteriorates (Section~\ref{sec:failure}).
By evaluating saved model checkpoints with and without the same
fixed harness, we distinguish two sources of decline:
the task-performing model loses ability, and the unchanged harness
becomes more harmful as the model changes.
In one task, a later model succeeds in 14 turns without a harness
but reaches the 50-turn limit under the old harness (Table~\ref{tab:diagnostic}b).
These findings motivate coordinating the two learners throughout
training, rather than relying on alternation alone.

We introduce \method{}, to our knowledge \textbf{the first framework}
to jointly train a harness-generation policy and a
task-execution policy through reinforcement learning.
We call the two policies the \textbf{Engineer} and the \textbf{Solver}.
The Engineer generates an executable harness, and the Solver
interacts with the environment under that harness.
The generated program and its multi-turn execution form a paired
training example (Figure~\ref{fig:overview}).
The shared task outcome trains both policies,
each on its own generated tokens.
Therefore, \textbf{JoMoHa implements RSI}: the improvement
policy itself learns from execution feedback, alongside the
task-performing policy.
We release our code at
\url{https://github.com/xudongwu-0/coevolve}.

We evaluate \method{} on ALFWorld, WebShop, and DBBench,
where it achieves the highest success rates among the evaluated
methods: 97.0\%, 45.8\%, and 68.0\%, respectively
(Section~\ref{sec:performance}, Table~\ref{tab:main}).

To coordinate the Engineer and Solver, \method{} combines
collaboration-aware value estimation, role-specific return
normalization, partner-trajectory weighting, and asymmetric
update schedules.
We examine these design choices through component studies
in Section~\ref{sec:ablations}.

Our contributions are threefold:
\begin{enumerate}
  \item \textbf{Diagnosing failure in alternating training.}
  Fixed-component tests distinguish declining Solver capability
  from worsening harness compatibility.
  Trajectory analysis reveals how incorrect harness state
  disrupts otherwise successful execution.
  Together, these findings highlight risks to both task
  execution and continued learning, motivating
  compatibility-aware joint training.

  \item \textbf{Joint model--harness reinforcement learning.}
  We introduce, to our knowledge, the first framework
  to jointly train an executable-harness generation policy
  and a task-execution policy through RL.
  The Engineer and Solver learn from the same task outcome,
  making both harness construction and task execution
  trainable within a shared learning process.

  \item \textbf{Coordination adaptations and empirical evaluation.}
  We adapt cooperative-learning mechanisms to the
  Engineer--Solver setting and evaluate \method{}
  across three environments.
  Frozen-Engineer controls, fixed-Solver harness comparisons,
  and component studies assess the benefits of training
  the Engineer, improving the harness, and coordinating
  the two policies.
\end{enumerate}

\section{Related Work}
\label{sec:related}

\paragraph{Model-weight optimization.}
Reinforcement learning improves language-model reasoning and
decision-making from task feedback \citep{shao2024deepseekmath}.
Single-Rollout Asynchronous Optimization (SAO) uses one rollout per
prompt, a learned value model, and token-level calibration to stabilize
asynchronous agentic RL \citep{hou2026sao}.
For tool-using agents, Search-R1 learns when and how to retrieve
information \citep{jin2025searchr1}, while ReTool learns to interleave
reasoning with code execution \citep{feng2025retool}.
RAGEN studies the training dynamics of multi-turn agents
\citep{wang2025ragen}.
These methods improve the task-performing policy under a prescribed
runtime, rather than jointly learning a separate policy that
constructs that runtime.
\method{} trains the Solver alongside an Engineer whose generated
harness shapes the Solver's execution.

\paragraph{Harness optimization.}
A complementary line improves the system surrounding the model.
ReAct organizes reasoning and action, while Self-Refine and Reflexion
use textual feedback to revise behavior
\citep{yao2023react,madaan2023selfrefine,shinn2023reflexion}.
OPRO and DSPy support instruction and pipeline optimization
\citep{yang2023opro,khattab2023dspy}.
ADAS and AFlow search over agent architectures and executable
workflows \citep{hu2024adas,zhang2024aflow}, and Meta-Harness
searches executable harnesses using prior code, scores, and
execution traces \citep{lee2026metaharness}.
These approaches motivate treating the runtime as an optimization
object, but improving a program does not necessarily train
its generator.
Harness-R1 makes runtime editing a learned capability by training
an Engineer through RL against a frozen target agent
\citep{shao2026harnessr1}.
\method{} further trains the Solver alongside the Engineer,
allowing both policies to adapt to their changing partner.

\paragraph{Joint learning and coordination.}
WHALE jointly improves model weights and harnesses through
alternating updates \citep{kim2026whale}.
Its published instantiation combines online RSFT with
Meta-Harness search driven by a fixed external proprietary
frontier model.
Compared with WHALE, \method{} learns the harness-generating
policy itself and trains both policies through RL from the
same executed program--rollout pair (Table~\ref{tab:related}).

Cooperative RL provides foundations for learning from team outcomes
\citep{yu2022mappo,foerster2017coma,kuba2021happo,
foerster2017replay,vanhasselt2016popart}.
These classical methods were primarily developed and evaluated
in non-language environments.
Adapting these ideas to LLM training, particularly our
program-mediated setting, is nontrivial:
the Engineer and Solver have separate token histories and different
decision horizons, while the Engineer's program shapes the Solver's
entire multi-turn rollout.
Their execution conditions also change as both policies learn.
Our adaptations address critic information, role-specific return
scales, partner-trajectory weighting, and update frequency.
Component studies examine their contributions in this setting,
and the complete coordination design improves performance over
the Joint-only baseline (Section~\ref{sec:ablations}).

\begin{table}[t]
\caption{\textbf{Comparison of model--harness learning procedures.}
Engineer and Solver columns describe parameter updates.
A fixed Engineer can still search for new harnesses.
For Harness-R1, the Solver is frozen during Engineer training.
A dash denotes no separate harness Engineer.}
\label{tab:related}
\centering
\small
\setlength{\tabcolsep}{3pt}
\renewcommand{\arraystretch}{1.12}
\begin{tabular}{@{}
  >{\raggedright\arraybackslash}p{0.29\columnwidth}
  >{\raggedright\arraybackslash}p{0.29\columnwidth}
  >{\raggedright\arraybackslash}p{0.17\columnwidth}
  >{\raggedright\arraybackslash}p{0.17\columnwidth}
@{}}
\toprule
Method & Harness adaptation & Engineer & Solver\\
\midrule
Search-R1 / ReTool & Prescribed        & --    & RL\\
Meta-Harness      & Code search       & External & Fixed\\
Harness-R1        & Learned edits     & RL    & Fixed\\
WHALE             & Code search       & External & RSFT\\
\textbf{\method{}}
& \textbf{Learned generation}
& \textbf{RL}
& \textbf{RL}\\
\bottomrule
\end{tabular}
\end{table}

\section{Background}
\label{sec:background}

Reinforcement learning (RL) improves a policy, a distribution over
actions, through interaction and reward \citep{sutton2018rl}.
A Markov decision process (MDP) describes this interaction using
states, actions, transition dynamics, and rewards: the next state
depends on the current state and action.
An episode is one task attempt.
Language agents observe only part of the environment, so their
policies use the available history.
An \emph{actor} is the trainable policy; a \emph{critic} predicts its
future return and provides a baseline for advantages---whether an
outcome is better or worse than expected.

For a task \(x\), the Engineer uses evidence \(e\) to sample a proposal
trajectory \(\tau_\engineer\).
Validation and fallback map it to the executed harness
\(h=\mathcal V(\tau_\engineer)\).
The Solver acts under \(h\), producing trajectory \(\tau_\solver\)
and terminal task reward \(R(x,\tau_\solver)\).
For role \(i\in\{\engineer,\solver\}\), \(\theta_i\) and \(\phi_i\)
are actor and critic parameters, with actor policy \(\pi_{\theta_i}\).
Indices \(t\), \(b\), and \(k\) denote generated-token positions,
paired episodes, and saved checkpoints, respectively;
\(\theta_\solver^{(k)}\) and \(h_k\) identify saved weights and code.
The no-harness condition \(\noharness\) retains the base executor.

SAO learns from one rollout per prompt using a critic
\citep{hou2026sao}.
Both roles share the task reward, while their returns also include
role-specific KL penalties that restrain divergence from fixed
reference policies.
% \textbf{Complete notation is collected in Appendix~\ref{app:notation}},
% Table~\ref{tab:notation};other main-text quantities are defined
% where used, and formal MDP and update-level details remain in the appendix.

\section{Why Learn Weights and Harnesses Together?}
\label{sec:problem}
\label{sec:findings}

We examine an alternating-training baseline that updates
the Solver and Engineer in separate phases.
Such training may lead to over-adaptation between the
Solver and its harness: each component adapts to a counterpart
that subsequently changes, potentially making their learned
behavior less compatible.
On ALFWorld, this baseline initially improves from 83.0\%
to 90.5\%, but then deteriorates across successive stages
to 31.5\% (Table~\ref{tab:alternating}).
In comparison, \method{} reaches 97.0\%.
We analyze the alternating run to investigate how harness
mismatch can affect both task execution and Solver learning.

\begin{table}[t]
\caption{\textbf{Performance degradation during alternating
training on ALFWorld.}
Each alternating stage uses its corresponding Solver--harness
pair \((\theta_\solver^{(k)},h_k)\).
The highest alternating-stage success and the result of
\method{} are bolded.
The fixed-code diagnosis is reported in Table~\ref{tab:diagnostic}.}
\label{tab:alternating}
\centering
\small
\setlength{\tabcolsep}{4pt}
\renewcommand{\arraystretch}{1.12}
\begin{tabular}{@{}lcr@{}}
\toprule
Training scheme & Checkpoint & Success (\%)\\
\midrule
\multirow{5}{*}{\shortstack[l]{Alternating\\training}}
& \(\theta_\solver^{(0)},h_0\) & 83.0\\
& \(\theta_\solver^{(1)},h_1\) & \textbf{90.5}\\
& \(\theta_\solver^{(2)},h_2\) & 78.5\\
& \(\theta_\solver^{(3)},h_3\) & 48.5\\
& \(\theta_\solver^{(4)},h_4\) & 31.5\\
\midrule
\textbf{\method{}}
& 144 rounds & \textbf{97.0}\\
\bottomrule
\end{tabular}
\end{table}

\subsection{Diagnosing Model--Harness Compatibility}
\label{sec:failure}

The stage-level results alone cannot distinguish declining
Solver ability from worsening harness compatibility,
because both components change across stages.
We therefore hold one harness fixed and evaluate it with
different Solver checkpoints.

The percentage success \(\mathrm{SR}(\theta_\solver,h)\) evaluates
fixed Solver weights under a saved harness.
The no-harness condition \(h=\varnothing\) removes the learned
harness while retaining the base executor.
The harness contribution to a fixed Solver is
\begin{equation}
\Delta_h(\theta_\solver,h)
=\mathrm{SR}(\theta_\solver,h)-\mathrm{SR}(\theta_\solver,\noharness).
\label{eq:patchgain}
\end{equation}
This comparison separates Solver performance without the
learned harness from the harness's immediate effect on execution.

Specifically, we select earlier Solver weights \(\theta_\solver^{(2)}\),
later weights \(\theta_\solver^{(3)}\), and the saved harness \(h_2\) from the
alternating run.
For each Solver, we evaluate the same fixed harness and
the no-harness condition on the same ALFWorld evaluation tasks.
The following aggregate comparisons and trajectory analyses
examine both execution interference and Solver degradation.

\begin{table*}[t]
\caption{\textbf{Execution interference and Solver degradation
during alternating training.}
(a) Aggregate success rates; harness effect is the difference
within each row in percentage points.
(b) Case 1 from the ALFWorld trajectory analysis:
outcomes and episode lengths across four Solver--harness
combinations.
Both \(h_2\) conditions execute the identical saved harness.
No harness retains the base executor but removes the
learned harness.}
\label{tab:diagnostic}
\centering
\small
\begin{minipage}[t]{0.51\textwidth}
\centering
\textbf{(a) Aggregate success (\%)}\par\medskip
\begin{tabular}{lrrr}
\toprule
Solver & No harness & Fixed \(h_2\) & Harness effect\\
\midrule
Earlier \(\theta_\solver^{(2)}\) & 82.5 & 78.5 & \(-4.0\)\\
Later \(\theta_\solver^{(3)}\)   & 48.0 & 31.0 & \(-17.0\)\\
\midrule
Change & \(-34.5\) & \(-47.5\) & \(-13.0\)\\
\bottomrule
\end{tabular}
\end{minipage}\hfill
\begin{minipage}[t]{0.46\textwidth}
\centering
\textbf{(b) Case 1: Clean sponge and place on cart}\par\medskip
\begin{tabular}{lcc}
\toprule
Solver & No harness & Fixed \(h_2\)\\
\midrule
Earlier \(\theta_\solver^{(2)}\)
& \shortstack{Success\\16 turns}
& \shortstack{Success\\14 turns}\\
\addlinespace
Later \(\theta_\solver^{(3)}\)
& \shortstack{Success\\14 turns}
& \shortstack{Timeout\\50 turns}\\
\bottomrule
\end{tabular}
\end{minipage}
\end{table*}

\paragraph{Separating execution interference from Solver degradation.}
Table~\ref{tab:diagnostic} reveals two distinct changes.
Across each row, adding the same harness reduces success
by 4.0 points for \(\theta_\solver^{(2)}\) but 17.0 points for
\(\theta_\solver^{(3)}\):
its negative effect increases by 13.0 points.
Down the no-harness column, success falls from 82.5\% to 48.0\%:
the Solver loses 34.5 points even when the learned harness
is removed.
Together, these account for the decline under the fixed harness:
\begin{equation}
\underbrace{-47.5}_{\text{system decline}}
=
\underbrace{-34.5}_{\text{Solver decline}}
+
\underbrace{-13.0}_{\text{compatibility loss}}.
\end{equation}

The first comparison shows that an unchanged harness can
increasingly interfere with execution as the Solver changes.
The additional loss cannot be explained by the Engineer
generating a worse new harness.
At \(\theta_\solver^{(3)}\), adding \(h_2\) changes success to failure on
18.0\% of evaluation tasks and failure to success on 1.0\%.

The second comparison shows that removing the harness does
not restore the later Solver's earlier performance.
The decline therefore extends beyond interference during
the current execution.
It is consistent with the concern that training under
mismatched harnesses may steer the Solver toward
less effective behavior.
These findings motivate maintaining Solver--harness
compatibility throughout training, rather than treating
the harness only as an inference-time aid.

\paragraph{Case 1: Harness-induced failure on ALFWorld.}
We examine evaluation trajectories produced by the earlier
and later Solver checkpoints from the alternating training
run analyzed above.
The task requires cleaning a sponge and placing it on a cart.
The later Solver succeeds in 14 turns without the learned
harness, but reaches the 50-turn limit under the old
harness \(h_2\) (Table~\ref{tab:diagnostic}b).
The earlier Solver succeeds in both conditions, taking
16 and 14 turns, respectively.
The later Solver therefore retains the behavior needed to
complete this task; the failure appears when it is combined
with the old harness.

The trajectory analysis reveals the following failure sequence:
\begin{enumerate}
  \item \textbf{The environment advances.}
  Cleaning has completed, so the remaining work is to finish
  navigation and placement.

  \item \textbf{The harness fails to update its state.}
  The executed action is stored in a nested field, but the old
  lookup uses a literal dotted key.
  The correct access is on the first line below;
  the old lookup is on the second:
\begin{verbatim}
ctx["action"]["final_action"]
ctx.get("action.final_action")
\end{verbatim}
  A dotted string is a different dictionary key, not a nested
  lookup, so the completed action is missed.
  A second state update can also overwrite earlier inventory
  evidence when the current observation omits the word
  indicating cleaning.

  \item \textbf{Stale guidance sends execution backward.}
  Although the environment has confirmed cleaning, the harness
  continues directing the Solver to find and clean the object.
  The episode loops through navigation and reaches the turn limit.
\end{enumerate}

\paragraph{Case 2: Solver failure without the learned harness.}
A contrasting ALFWorld case from the same training run
exposes a failure in the no-harness condition.
After cleaning a potato and navigating to the destination,
the later Solver omits the final placement action and
eventually reaches the turn limit.
Because no learned harness is active, this failure cannot
be attributed to interference from \(h_2\).
It illustrates why removing an incompatible harness alone
cannot address all failures of the trained Solver.
% Appendix~\ref{app:cases} provides the corresponding trajectories
% and additional cases;
% Appendix~\ref{app:interventions} reports separate manual
% harness interventions.

\subsection{From Compatibility to Coordinated Learning}

Taken together, these findings suggest that \textbf{a mismatched
harness can impair current task execution and may also
distort the experience from which the Solver learns}.
This creates a risk of a negative feedback loop in which
ineffective harness guidance and declining Solver capability
reinforce one another.
Maintaining a compatible Solver--harness pair throughout
learning is therefore important.

Our unit of learning is a complete task attempt:
the generated harness, the actions taken under it,
and the final outcome.
The Engineer learns to generate effective harnesses,
while the Solver learns to complete tasks under them.
To mitigate this risk, our method uses their shared task
outcome to coordinate value estimation and policy updates,
supporting the joint adaptation of both policies.

\section{\method{}}
\label{sec:method}

\method{} jointly trains two LLM policies:
an Engineer that generates a harness and a Solver that
performs tasks under it.
Both learn from the outcome of the same task attempt
(Figure~\ref{fig:overview}).
To coordinate their learning, we use collaboration-aware
value estimation, role-specific return normalization,
partner-trajectory weighting, and asymmetric update schedules.

\subsection{Learning from a Shared Task Outcome}

For a task \(x\), the Engineer receives historical feedback
\(e\) and samples a proposal trajectory \(\tau_\engineer\).
Validation and fallback map that proposal to the executed harness
\(h=\mathcal V(\tau_\engineer)\).
The Solver then interacts with the environment under \(h\),
producing trajectory \(\tau_\solver\) and task reward
\(R(x,\tau_\solver)\).
An invalid proposal maps to \(h=\noharness\), retaining the base executor.

The two policies share the objective
\begin{equation}
J(\theta_\engineer,\theta_\solver)
=
\E_{x,e,h,\tau_\solver}\left[R(x,\tau_\solver)\right],
\label{eq:objective}
\end{equation}
where \(h\) is the validation/fallback result and \(\tau_\solver\)
is collected by the Solver under that harness.
The task reward is binary on ALFWorld.

The same reward trains both policies:
the Engineer learns which harnesses support successful
execution, while the Solver learns how to complete tasks
under those harnesses.
Each policy is updated on its own generated outputs,
without differentiating through harness execution.
We use a KL-regularized SAO-based actor--critic backbone
\citep{hou2026sao}, with a fixed reference policy for each role.
The local realization retains behavior--proximal token-ratio
decoupling% , as detailed in Appendix~\ref{app:ownpolicy}
.
% The detailed training objectives and treatment of invalid
% harnesses are given in
% Appendices~\ref{app:learningguide} and~\ref{app:credit}.

\subsection{Value Estimation for Two Different Roles}
\label{sec:context}

Each policy has an independent critic that estimates its
expected return, including task rewards and its own KL penalty.
The critics provide baselines for computing advantages:
whether an outcome is better or worse than expected.

\paragraph{Context: account for collaboration conditions.}
The usefulness of an Engineer proposal depends on the Solver,
while the Solver's performance depends on the harness.
A critic \(V_{\phi_i}(z_{i,t},c_i)\) predicts the role's expected
remaining regularized return from its current history \(z_{i,t}\)
and a short collaboration summary \(c_i\).
The supplemental fields describe the task identifier, training version,
and role-specific execution conditions.
The task field is a path or identifier, not a full environment observation.
Context is provided only to the critics and does not change
the actors' inputs.
% Appendix~\ref{app:context} specifies its contents.

\paragraph{ValueNorm: account for different return scales.}
The Engineer generates a harness once, whereas the Solver
acts over multiple turns.
Their different output lengths and KL penalties can produce
different return scales despite the shared task reward.
Writing the raw, unweighted critic target as \(G_{i,t}\), we normalize critic targets
separately for each role:
\begin{align}
\widehat G_{i,t}
&=
\frac{G_{i,t}-m_i}{\sigma_i},
\nonumber\\
V_{\phi_i}(z_{i,t},c_i)
&=
\sigma_i\widehat V_{\phi_i}(z_{i,t},c_i)+m_i,
\label{eq:valuenorm}
\end{align}
where \(m_i\) and \(\sigma_i>0\) are running return statistics.
Values are restored to their original scale before
advantages are computed.
Hats denote normalized targets and predictions.
This critic-target normalization is separate from
standardizing actor advantages.
% The raw GAE target is computed before advantage standardization or
% Peer weighting, as defined in Appendix~\ref{app:advantages}.

\subsection{Coordinating Policy Updates}
\label{sec:peer}

We compute advantages using generalized advantage estimation
\citep{schulman2016gae} and standardize them separately
for the two policies.
% The SAO-based path uses the local token-ratio calibration described in
% Appendix~\ref{app:ownpolicy}.
We introduce two additional mechanisms to account for
the changing partner and the different adaptation needs
of the two roles.

\paragraph{Peer: weight experience using the partner's trajectory.}
As both policies change, a recorded task attempt may become
more or less representative of their current collaboration.
Peer adjusts its influence using the paired partner trajectory:
\begin{equation}
\widetilde A_{\engineer,b,t}
=
w_{\solver,b}\bar A_{\engineer,b,t},
\qquad
\widetilde A_{\solver,b,t}
=
w_{\engineer,b}\bar A_{\solver,b,t},
\end{equation}
where \(\bar A_{i,b,t}\) is the role-standardized advantage.
Each partner weight compares the likelihood of the recorded
trajectory under the policy snapshot used for the current
update with its likelihood when collected.
We bound the weight to \([0.5,2]\), treat it as a constant
during optimization, and leave critic targets unchanged.
% Peer is an empirical weighting rule;
% Appendix~\ref{app:peerweighting} gives its calculation.

\paragraph{Schedule: give the Solver additional adaptation.}
\label{sec:schedule}
Changing the Engineer changes the harnesses under which
the Solver learns.
To give the Solver additional adaptation without another
Engineer update, the coordination study repeats
\emph{joint, joint, Solver-only} rounds.
The resulting update ratio is
Engineer:Solver \(=2:3\).

Solver-only rounds freeze the Engineer's parameters but
still sample fresh harnesses.
They use unit partner weights.
Thus, the Solver continues learning under sampled harnesses
while their generating policy remains fixed.
% Warmup settings and the detailed update order are provided
% in Appendix~\ref{app:algorithm}.

Algorithm~\ref{alg:learning} summarizes the procedure.
At deployment, we select a saved harness and evaluate it
with a frozen Solver checkpoint.

\begin{algorithm}[!ht]
\caption{Joint learning of the Engineer and Solver}
\label{alg:learning}
\begin{algorithmic}[1]
\REQUIRE Engineer and Solver policies, independent critics,
reference policies, task/evidence stream
\FOR{each training round}
  \STATE Choose active roles according to Schedule.
  \STATE Engineer samples a code proposal from historical feedback.
  \STATE Validate it to obtain the executed harness and run the Solver.
  \STATE Record both outputs and their shared task reward.
  \FOR{each active role}
    \STATE Estimate values using Context and ValueNorm.
    \STATE Compute and standardize advantages.
    \STATE Apply Peer weights and update the actor through the SAO-based path.
    \STATE Update the critic on normalized return targets.
  \ENDFOR
\ENDFOR
\STATE Select a saved harness and Solver checkpoint for evaluation.
\end{algorithmic}
\end{algorithm}

\begin{table*}[t]
\caption{\textbf{Benchmark comparison across agent and harness-optimization methods.}
Frontier models serve as fixed harness editors.
Reflection uses success@2 (\(\dagger\)).
Avg.\ is the mean of ALFWorld All, WebShop success, and DBBench success;
it excludes WebShop score.
Best and second-best distinct values are bold on red and underlined on blue,
including ties.}

% Confidence intervals are collected in Appendix~\ref{app:statistics}.

\label{tab:main}
\centering
\small
\setlength{\tabcolsep}{3.2pt}
\begin{tabular}{@{}>{\raggedright\arraybackslash}p{0.215\textwidth}*{11}{r}@{}}
\toprule
\multirow{2}{*}{Method} & \multicolumn{7}{c}{ALFWorld} &
\multicolumn{2}{c}{WebShop} & DBBench & \multirow{2}{*}{Avg.}\\
\cmidrule(lr){2-8}\cmidrule(lr){9-10}\cmidrule(lr){11-11}
& Pick & Look & Clean & Heat & Cool & Pick2 & All & Score & Succ. & Succ. &\\
\midrule
Qwen3.5-9B (with default harness) &
75.8 & 66.7 & 16.9 & 28.8 & 9.2 & 47.5 & 40.6 & 66.0 & 31.2 & 61.0 & 44.3\\
\addlinespace
\multicolumn{12}{@{}l}{\textbf{Prompt-based Agentic Methods}}\\
ReAct &
79.8 & 54.8 & 27.3 & 27.4 & 10.3 & 53.3 & 43.4 & 66.8 & 37.4 & 61.7 & 47.5\\
Self-Refine &
70.7 & 45.2 & 11.7 & 28.8 & 6.9 & 57.4 & 39.0 & 61.7 & 29.0 & 57.3 & 41.8\\
Reflection\(^\dagger\) &
91.9 & 90.5 & 24.7 & 56.2 & 19.5 & 73.8 & 59.2 & 61.2 & \second{43.6} & 64.7 & 55.8\\
\addlinespace
\multicolumn{12}{@{}l}{\textbf{Fixed Frontier Harness Editors}}\\
Qwen3.5-397B &
76.8 & 66.7 & 14.3 & 31.5 & 11.5 & 47.5 & 41.2 & 66.4 & 32.8 & 63.3 & 45.8\\
GLM-5.2 &
77.8 & 78.6 & 18.2 & 27.4 & 16.1 & 54.9 & 45.0 & 68.4 & 36.0 & 65.3 & 48.8\\
Kimi-K2.6 &
73.7 & 71.4 & 14.3 & 30.1 & 12.6 & 49.2 & 41.4 & 63.7 & 31.8 & 62.7 & 45.3\\
DeepSeek-V4-Pro &
72.7 & 69.0 & 13.0 & 30.1 & 16.1 & 47.5 & 41.0 & 65.1 & 32.4 & 64.3 & 45.9\\
Gemini-3.5-Flash &
67.7 & 69.0 & 14.3 & 24.7 & 10.3 & 35.2 & 35.4 & 63.2 & 33.6 & 64.0 & 44.3\\
GPT-5.5 &
80.8 & 78.6 & 31.2 & 28.8 & 17.2 & 35.2 & 43.2 & 61.4 & 36.6 & 64.0 & 47.9\\
\addlinespace
\multicolumn{12}{@{}l}{\textbf{Harness-Adapted Agents}}\\
Agent SFT + Harness-R1 &
93.9 & \best{100.0} & \second{81.8} & 74.0 & \second{72.4} & 86.1 & \second{84.0} &
68.7 & 43.0 & \second{65.7} & \second{64.2}\\
WHALE &
\second{97.7} & \second{92.9} & 55.2 & \second{79.3} & 66.7 & \second{87.8} &
80.5 & \second{71.63} & 43.4 & 65.3 & 63.1\\
\textbf{\method{} (ours)} &
\best{100.0} & \best{100.0} & \best{100.0} & \best{96.6} &
\best{97.2} & \best{91.8} &
\best{97.0} & \best{73.9} & \best{45.8} & \best{68.0} & \best{70.3}\\
\bottomrule
\end{tabular}
\end{table*}

\section{Experiments}
\label{sec:setup}
\label{sec:results}

\paragraph{Experimental setup.}
We evaluate \method{} on ALFWorld, WebShop, and DBBench, which test
household task completion, goal-directed shopping, and database interaction,
respectively \citep{shridhar2021alfworld,yao2022webshop,liu2024agentbench}.
We report success rates in all three environments, a separate WebShop score,
and the equally weighted mean of the three success rates.
The ALFWorld coordination study uses separate Qwen3.5-9B-class Engineer
and Solver actors with independent critics, initialized from Agent SFT,
and collects eight paired episodes per round for 144 rounds.
% Training details and confidence intervals are provided in
% Appendices~\ref{app:protocol} and~\ref{app:statistics}.

\paragraph{Baselines.}
Except for locally evaluated \method{} and WHALE, scores are taken from
Harness-R1 \citep{shao2026harnessr1}; we use its training and test sets,
with checkpoint and harness selection on a separate validation set.
The task model is Qwen3.5-9B; frontier model names denote harness editors,
not replacement Solvers.
Agent SFT + Harness-R1 additionally fine-tunes the task model,
while WHALE uses its RSFT-plus-search recipe.
WHALE alternates after eight environment interactions (16 training steps),
uses the same harness-search settings as \method{}, and matches its
total environment-interaction budget.
Section~\ref{sec:ablations} examines one-sided learning, frozen-Engineer
training, program swaps, and coordination components.

\subsection{Main Results}
\label{sec:performance}

\method{} achieves the highest success rates among the evaluated methods
on all three benchmarks: 97.0\% on ALFWorld, 45.8\% on WebShop, and
68.0\% on DBBench (Table~\ref{tab:main}).
Its average success is 70.3\%, exceeding Agent SFT + Harness-R1 by
6.1 percentage points and WHALE by 7.2 points.
The gains over WHALE are 16.5, 2.4, and 2.7 points across the three
environments, respectively.

The ALFWorld breakdown locates the largest gains on tasks that transform
an object's state before placement.
Against the strongest competing result in each category, \method{} gains
18.2 points on Clean, 17.3 on Heat, and 24.8 on Cool, compared with
0.0--4.0 points on the remaining categories.
This pattern extends beyond differences in ceiling effects: on Heat and
Cool, the gains remove approximately 83.6\% and 89.9\% of the respective
best competitor's errors
% (Appendix~\ref{app:categories})
.
These tasks require execution to distinguish a completed transformation
from the remaining navigation and placement, making the results consistent
with the state-tracking issue diagnosed in Section~\ref{sec:failure}.
Pick2 remains the weakest category at 91.8\%, indicating remaining room
for improvement when multiple objects must be collected and placed.

\begin{table}[t]
\caption{\textbf{Harness utility at fixed Solver weights.}
Fixed-program cross-evaluation: columns fix Solvers and rows fix programs.
Earlier and later denote programs and Solvers saved at checkpoints 48
and 144, respectively.
The initial program comes from the initial Engineer.
Colors follow Table~\ref{tab:main}.}

\label{tab:fixed}
\centering
\small
\begin{tabular}{lrr}
\toprule
Harness & Earlier Solver & Later Solver\\
\midrule
No harness & 90.0 & 91.5\\
Initial program & \second{91.5} & 92.5\\
Earlier program & \best{95.0} & \second{95.0}\\
Later program & \best{95.0} & \best{96.5}\\
\bottomrule
\end{tabular}
\end{table}

\begin{table}[t]
\caption{\textbf{Coordination-component additions.}
Success uses the saved harness and Solver.
Each addition row changes one coordination component relative to Joint-only;
the full method combines all four.
Best and second best follow Table~\ref{tab:main}.}
\label{tab:components}
\centering
\small
\begin{tabular}{lr}
\toprule
Variant & Success (\%)\\
\midrule
Joint-only & 88.5\\
\quad + Schedule & 92.0\\
\quad + Context & \second{92.5}\\
\quad + Peer & 92.0\\
\quad + ValueNorm & 91.5\\
\method{} & \best{97.0}\\
\bottomrule
\end{tabular}
\end{table}

% Intervals are reported in Appendix~\ref{app:componentcurves}.

\subsection{Ablation Studies}
\label{sec:ablations}

We examine three sources of improvement: training the Engineer rather than
only resampling its outputs, improving the harness at fixed Solver weights,
and coordinating the two policies beyond the Joint-only baseline.

\paragraph{Training the Engineer beyond fresh harness sampling.}
Frozen Engineer fixes the initial Engineer parameters while continuing to
sample fresh harnesses and train the Solver.
It reaches 90.5\% at step 144, versus 97.0\% for \method{}
(Figure~\ref{fig:curves}); even its best reported checkpoint, 91.5\%
at step 108, remains 5.5 points below the complete method.
Thus, the gap persists when the control can both resample programs and
use its best reported checkpoint.
Weight-only and Harness-only reach 89.5\% and 80.5\%, respectively.
% (Appendix~\ref{app:latest}).
Together, these controls favor training both policies over the evaluated
one-sided recipes, while the component study below examines the additional
benefit of coordination.

\paragraph{Harness utility at unchanged Solver weights.}
Learning curves alone cannot separate improvements in Solver ability from
improvements in the generated runtime.
We therefore cross-evaluate saved programs and Solver checkpoints
(Table~\ref{tab:fixed}).
At the later Solver, replacing the initial program with the later program
raises success from 92.5\% to 96.5\%, without changing model weights.
Relative to no harness, the later program gains 5.0 points, with no observed
success-to-failure regressions on this evaluation set.
% ; its paired interval
% is reported in Appendix~\ref{app:fixed}.
The earlier and later learned programs also both improve the earlier Solver
from 91.5\% under the initial program to 95.0\%.
These results show that the saved programs contribute useful behavior
beyond the initial runtime, including at a different Solver checkpoint.
The equal 95.0\% scores on the earlier Solver, but 95.0\% versus 96.5\%
on the later Solver, further show that program value depends on the
Solver with which it is paired.

\paragraph{Coordinating the two learners.}
Joint-only trains both policies from paired task outcomes without Context,
ValueNorm, Peer, or Schedule.
Adding each mechanism separately raises its final success from 88.5\%
to 91.5--92.5\%; combining all four reaches 97.0\%, an 8.5-point
improvement over Joint-only (Table~\ref{tab:components}).
Context yields the highest single-addition point estimate, 92.5\%,
consistent with the value of informing each critic about its collaboration
conditions.
Schedule and Peer both achieve 92.0\%, while ValueNorm achieves 91.5\%. 
% see Appendix~\ref{app:componentcurves} for a detailed analysis

\section{Discussion and Conclusion}
\label{sec:conclusion}

One model learns to write the runtime; another learns to act under it.
\method{} trains both from the same task result while keeping their token
histories and value estimates separate.
Fixed-component tests measure the two sides of this interaction.
At fixed Solver weights, saved programs improve on the initial code.
Learning the writer outperforms freezing it while still sampling new code,
and coordinating the two learners improves on joint updates alone.
The complete system has the highest reported success on all three
environments in the benchmark table, averaging 70.3\%.
This is RSI through execution feedback: both the task policy and the
policy that constructs its runtime are learned within the system.

% \section*{Impact Statement}

% This work studies how language-model agents can adapt both their policies
% and their executable runtime.
% The evaluation covers simulated household tasks, shopping, and database
% interaction.
% More adaptive harnesses could improve reliability, but generated code can
% also impose incorrect constraints or conceal state-tracking errors.
% Validation, constrained execution, and execution-level auditing remain
% important when applying such methods outside controlled environments.
% The present results do not establish safety for unrestricted tool use.

\label{end:mainbody}
\bibliography{references}
\bibliographystyle{icml2026}

% \clearpage
% \appendix
% \onecolumn
% \input{appendix}

\end{document}